\documentclass[sigconf,nonacm]{acmart}
\usepackage{xspace}
\usepackage{tabularx}
\usepackage{booktabs}
\usepackage{multirow}
\usepackage{xcolor}
\usepackage{comment}
\usepackage{amsmath}
\usepackage{algorithm}
\usepackage{algpseudocode}
\usepackage{verbatim}
\usepackage{tcolorbox}
\usepackage{tabularx}
\usepackage{enumitem}
\usepackage{subcaption}
\usepackage{booktabs}

\newcommand{\GALA}{\textsf{GALA}\xspace}
\newcommand{\TWIST}{\textsf{TWIST}\xspace}
\newcommand{\surescore}{\textsf{SURE-Score}\xspace}

\AtBeginDocument{%
  }

\begin{document}
\emergencystretch 3em

\title{GALA: Can Graph-Augmented Large Language Model Agentic Workflows Elevate Root Cause Analysis?}

\author{Yifang Tian}
\email{yifang.tian@mail.utoronto.ca}
\affiliation{%
  \institution{University of Toronto}
  \city{Toronto}
  \state{Ontario}
  \country{Canada}
}

\author{Yaming Liu}
\email{yaming.liu@mail.utoronto.ca}
\affiliation{%
  \institution{University of Toronto}
  \city{Toronto}
  \state{Ontario}
  \country{Canada}
}

\author{Zichun Chong}
\email{zichun.chong@mail.utoronto.ca}
\affiliation{%
  \institution{University of Toronto}
  \city{Toronto}
  \state{Ontario}
  \country{Canada}
}

\author{Zihang Huang}
\email{zihang.huang@mail.utoronto.ca}
\affiliation{%
  \institution{University of Toronto}
  \city{Toronto}
  \state{Ontario}
  \country{Canada}
}

\author{Hans-Arno Jacobsen}
\email{jacobsen@eecg.toronto.edu}
\affiliation{%
  \institution{University of Toronto}
  \city{Toronto}
  \state{Ontario}
  \country{Canada}
}

\renewcommand{\shortauthors}{Yifang et al.}

\begin{abstract}
Root cause analysis (RCA) in microservice systems is challenging, requiring on-call engineers to rapidly diagnose failures across heterogeneous telemetry such as metrics, logs, and traces. Traditional RCA methods often focus on single modalities or merely rank suspect services, falling short of providing actionable diagnostic insights with remediation guidance. This paper introduces \GALA, a novel multi-modal framework that combines statistical causal inference with LLM-driven iterative reasoning for enhanced RCA. 
Evaluated on an open-source benchmark, \GALA achieves substantial improvements over state-of-the-art methods of up to 42.22\% accuracy. Our 
novel human-guided LLM evaluation score shows \GALA generates significantly more causally sound and actionable diagnostic outputs than existing methods. Through comprehensive experiments and a case study, we show that \GALA bridges the gap between automated failure diagnosis and practical incident resolution by providing both accurate root cause identification and human-interpretable remediation guidance.
\end{abstract}






\maketitle

\section{Introduction}

Modern software systems increasingly adopt microservice architectures to achieve scalability, maintainability, and deployment flexibility~\cite{dragoni2017microservices}. However, this architectural paradigm introduces significant operational complexity, as applications are decomposed into numerous loosely coupled services that communicate through intricate dependency graphs~\cite{dragoni2017microservices}. Root cause analysis (RCA) refers to the process of identifying the origin of failure after their occurrence. RCA in microservice environments becomes a challenging task that requires correlating symptoms across multiple services, analyzing heterogeneous telemetry data, and understanding complex inter-service dependencies.

Existing RCA approaches suffer from several limitations. Traditional statistical methods~\cite{lin2024root,pham2024baro,xin2023causalrca} often operate on single data modalities or use static heuristics that fail to capture the dynamic nature of microservice failures. 
For example, some failures occur only under rare or dynamic workloads—such as a memory leak triggered by a specific API sequence—and often escape statistical detection, as they do not produce persistent or recurring metric anomalies~\cite{xie2024cloud}. While some multi-modal approaches exist~\cite{fu2025msfsanomaly,yu2023nezha,zheng2024multi}, they typically collapse heterogeneous data into unified representations, potentially losing critical modality-specific information. Moreover, most existing methods focus solely on ranking potential root causes, providing limited explanatory context or actionable recommendations for incident resolution.

The emergence of Large Language Models (LLMs) has opened new possibilities for automating complex reasoning tasks~\cite{brown2020language}. Recent advances in LLM capabilities, particularly in structured reasoning with chain-of-thought prompting~\cite{wei2022chain} and agentic workflows~\cite{yao2023react}, have demonstrated their potential for multi-step problem-solving scenarios. However, applying LLMs to RCA presents unique challenges: the need to process multi-modal telemetry data, maintain causal reasoning consistency, and generate not just accurate fault rankings but also actionable remediation guidance.

We present \GALA (\textbf{G}raph-\textbf{A}ugmented Large \textbf{L}anguage Model \textbf{A}gentic Workflow), a novel framework that addresses these limitations. We also introduce \surescore, a human-guided evaluation framework specifically designed for assessing RCA-related natural language generation (NLG) tasks, addressing the inadequacy of traditional NLG metrics for operational contexts. We evaluate \GALA based on the RCAEval~\cite{pham2025benchmark} dataset, and our experimental results demonstrate significant improvements over state-of-the-art methods, achieving up to 42.22\% Top-1 accuracy and superior performance in incident summarization quality as measured by \surescore.
The main contribution of this paper are five-fold:

\begin{enumerate} [leftmargin=*]

\item \textbf{Multi-Modal Causal Inference Integration}: We develop a dual-modality hypothesis generation mechanism that combines metrics-based causal inference with a novel trace-based scoring approach \TWIST (\textbf{T}race‐based \textbf{W}eighted \textbf{I}mpact \textbf{S}coring \& \textbf{T}hresholding). This yields complementary initial root cause hypotheses that leverage both statistical causal discovery and graph-based trace analysis. 

\item \textbf{Modality-Preserving Diagnostic Synthesis}: We introduce a structured approach to preserve and present heterogeneous telemetry data without collapsing it into unified representations. This maintains critical modality-specific information while enabling effective LLM-based reasoning, resulting in superior causal soundness and incident specificity in our evaluations. 

\item \textbf{Iterative Agentic Reasoning Workflow}: We design a multi-agent LLM system where specialized agents collaborate to refine root cause hypotheses through evidence-based analysis and re-ranking. This iterative approach consistently improves ranking accuracy across all baseline methods.

\item \textbf{Integrated Incident Response Generation}: We develop a mechanism that automatically generates comprehensive incident summaries and prioritized action recommendations, bridging the gap between fault identification and practical remediation. This addresses the critical need for actionable guidance in production environments.

\item \textbf{Domain-Specific Evaluation Framework}: We introduce \surescore, a human-guided evaluation framework specifically designed for assessing RCA-related natural language generation tasks. This addresses the inadequacy of traditional NLG metrics for operational contexts and provides transparent, reproducible evaluation aligned with practitioner judgments.

\end{enumerate}

\section{Background}

In this section we provide an review of key concepts and abstractions that underpin the methodologies discussed in the subsequent sections of this paper. \textbf{Microservice} architectures decompose applications into small, loosely coupled, and independently deployable services, each encapsulating a distinct business capability~\cite{dragoni2017microservices}. A \textbf{pod} is the smallest deployable units for services, and serves as the fundamental unit of analysis in our work. 
\textbf{Telemetry data} which includes metrics, logs, and traces are critical observability artifacts that support monitoring and fault diagnosis. 
\textbf{Metrics} are structured, quantitative measurements collected at regular intervals (e.g., latency, error rates), offering a high-level view of system performance and health over time ~\cite{bhosale2022metrics}. 
\textbf{Logs} consist of timestamped, event‐level records emitted by individual services, enabling in‐depth, post‐hoc analysis of errors and anomalous behaviors~\cite{yu2023logreducer}.
\textbf{Traces} capture system call relationships and timing information across service boundaries. 
A \textbf{span} in tracing is the fundamental unit of work representing a single operation within a service. Each span is uniquely identified by a \texttt{spanID} and associated with a global \texttt{traceID}, and it records a timestamp, duration, and metadata such as service name. Spans also reference a \texttt{parentSpanID}, enabling the construction of \textbf{Trace DAG} (Directed Acyclic Graph). In Trace DAG, the complete request execution path across multiple services are modeled is a DAG where nodes represent individual spans and directed edges encode parent-child call relationships based on the \texttt{parentSpanID} references~\cite{sigelman2010dapper}.

\textbf{Root cause analysis (RCA)} is the systematic process of identifying the root cause of failures or anomalous behavior in complex systems. In the context of microservice-based system, RCA aims to investigate and correlate symptoms (e.g., increased latency, spikes in errors, or service outages) back to the lowest-level operational failures (e.g., memory leaks, network packet loss, configuration errors) that triggered the situation. Traditionally, RCA is performed manually by on-call engineers (OCEs) or site reliability engineers (SREs), who must sift through heterogeneous telemetry data (metrics, logs, and trace records), apply domain expertise, and iterate on various potential origins until the true cause is found.

\textbf{Agentic workflow} (e.g., ReAct~\cite{yao2023react}) represents a paradigm where LLMs~\cite{brown2020language} are integrated with external actions and iterative reasoning capabilities to perform complex problem-solving tasks. Unlike traditional LLM applications that reason over a single pass~\cite{wei2022chain,wang2023selfconsistency,yao2023tree}, agentic workflows enable LLMs to engage in multi-step reasoning processes, where the model can plan actions, execute them, observe outcomes, and adjust its approach accordingly. This paradigm closely resembles the human approach to problem solving, and especially the context of RCA, where agentic workflows are particularly valuable because fault diagnosis inherently requires iterative hypothesis formation, evidence gathering, and refinement. 
\textbf{Natural Language Generation (NLG)} refers to the automated production of fluent and contextually appropriate text from structured or unstructured data~\cite{gatt2018survey}. In the context of system monitoring and RCA, NLG is increasingly important for converting multi-modal telemetry into concise, human-interpretable incident summaries and practical remediation guidance~\cite{yu2023nezha,pei2025flow}.

\section{Problem Statement}
\label{sec:problem-statement}

\subsection{Motivating Incident Example}

Consider a cloud‐native retail application comprising multiple microservices (e.g., \texttt{Catalog Service}, \texttt{Payment Service}, \texttt{Frontend Service}), and at 04:12 UTC on June 1, an incident is detected: \textit{elevated end‐to‐end latency and intermittent errors in payment processing.}
A few minutes after the incident, an OCE is alerted, starts troubleshooting following best practices. The engineer first examines the metrics dashboard, which implicates an initial root cause hypothesis of \texttt{Redis Cache} as the core reason due to a transient memory spike in its timeseries. A subsequent manual inspection of Redis logs reveals no error entries, and trace data confirm normal  Remote Procedure Call (RPC) throughput, prompting a revision of the hypothesis. The engineer then evaluates a trace-only analysis that highlights \texttt{Frontend Service} as the most frequent node on anomalous call paths; however, \texttt{Frontend} logs are error-free and downstream services show no performance degradation. 

Only after correlating structured log exceptions and a sustained memory‐usage anomaly in \texttt{Catalog Service}, together with its central position in the service-call graph, can the true cause of memory leak be isolated. Throughout this process, the engineer continuously integrates multi-modal telemetry to form and validate root-cause hypotheses. More importantly, providing ranking alone by the diagnostic tool proves insufficient: on-call personnel also require concise incident summaries and prioritized action recommendations to expedite remediation, which motivates an agentic diagnostic system that not only produces accurate fault rankings but also automatically generates human-readable summaries and targeted recovery steps.

\subsection{Problem Definition}
\label{sec:problem-definition}

We formalize the RCA task in microservice systems as follows. Given a microservice deployment experiencing an incident, we observe telemetry data across three modalities—metrics, logs, and traces—collected during the incident time window. A root cause is defined as the fundamental failure that triggered the observed symptoms. This root cause corresponds to the specific service component and failure mode (e.g., memory leak in \texttt{Currency Service}) that, when addressed, resolves the incident. In our evaluation framework, ground truth root causes are established through controlled fault injection experiments where the injected failure location and type are known. The RCA task requires developing a method that effectively integrates heterogeneous telemetry signals to produce:

\begin{enumerate} [leftmargin=*]

    \item \textbf{Ranked Root Cause Identification}: A prioritized ranking of candidate services ordered by their likelihood of being the root cause, enabling rapid fault localization.

    \item  \textbf{Comprehensive Incident Response}: Human-readable incident summaries that explain the causal chain from symptoms to root cause, coupled with prioritized, actionable remediation recommendations.

\end{enumerate}

\section{The \GALA Design} 
\label{sec:methodology}

\begin{figure*}[h!]
    \centering
    \centerline{
        \includegraphics[width=\textwidth]{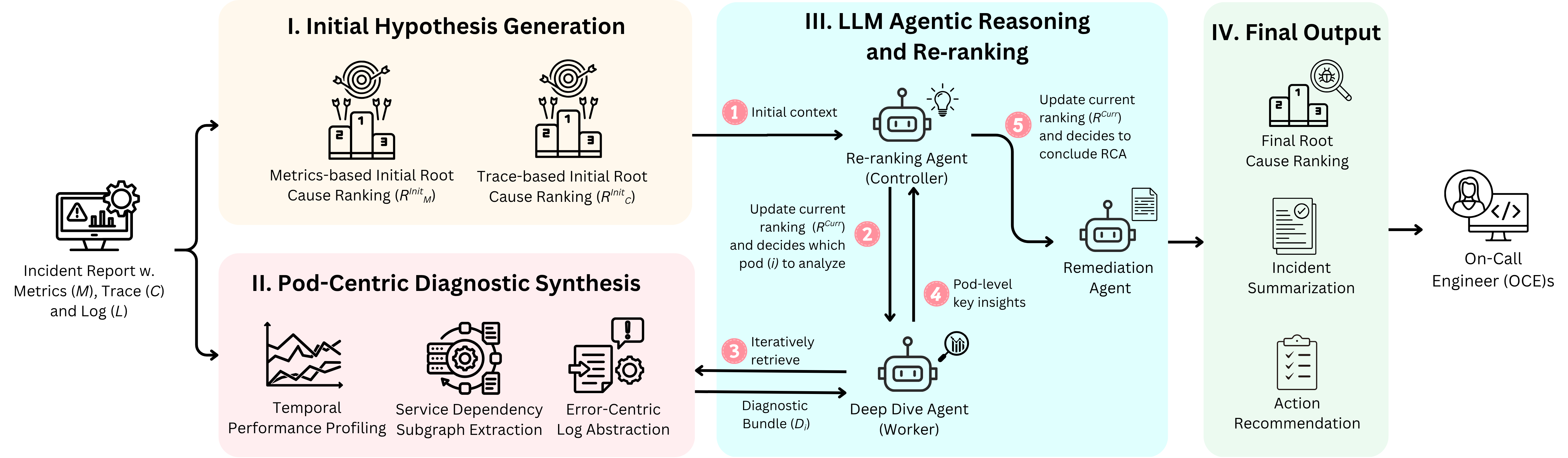}
    }
    \caption{Overview of \GALA. The framework employs an agentic, iterative workflow for RCA, organized into: (1) the Initial Hypothesis Generation phase, whose output is submitted to the Re‐ranking Agent to decide “Analyze Next” or “Finish”; (2) targeted pod‐level investigation delegated to the Deep Dive Analysis Agent; (3) iterative synthesis of Pod‐Centric Diagnostic artifacts into concise summaries; (4) combination of these summaries with trace‐graph predecessor/successor context to update the diagnostic report; and (5) repetition of steps (2–4) until conclusion or max iterations are reached, after which the Remediation Agent produces the incident summary and action recommendations.}

    \label{fig:GALA}
\end{figure*}

In this section, we describe our approach \GALA, a unified, multi-modal RCA framework for RCA in microservice-based systems. As illustrated in Figure~\ref{fig:GALA}, \GALA comprises four distinct yet interdependent phases: \textit{I. Initial Hypothesis Generation} (Section~\ref{Initial Hypothesis Generation}), \textit{II. Pod-Centric Diagnostic Synthesis} (Section~\ref{Pod-Centric Diagnostic Synthesis}), \textit{III. LLM Agentic Reasoning and Re‐ranking} (Section~\ref{LLM Agentic Reasoning and Re‐ranking}), and \textit{IV. Final Output Preparation} (Section~\ref{Final Output Preparation}). Each phase contributes a targeted transformation or reasoning step to convert raw telemetry into actionable incident remediation guidance. 

\GALA ingests the telemetry data in the incident report as input, which includes metrics, logs and traces across the microservice pod set \(\mathcal{P}\) within the incident time window. For each pod \(i\in\mathcal{P}\), we denote its metrics, logs, and traces by \(m_i\), \(l_i\), and \(c_i\), respectively,  where \(M\), \(L\), and \(C\) represent the complete collections of metrics, logs, and trace data across all pods in the system.  Phase I applies lightweight causal inference on metrics \(M=\{m_i\}\) to produce a metrics‐based initial ranking \(R^{\mathrm{Init}}_{M}\), and invokes our proposal of trace‐based module \TWIST  to yield \(R^{\mathrm{Init}}_{C}\). These two complementary lists form the initial hypotheses for Phase III’s LLM exploration. In Phase II, each root-cause candidate pod’s telemetry is distilled into a diagnostic bundle \(D_i\) that consists of three semantically rich artifacts: Temporal Performance Profiling, Service Dependency Subgraph Extraction and Error-Centric Log Abstraction to support LLM reasoning. These artifacts serve later as tailored prompts that preserve temporal trends, dependency structures, and failure semantics, respectively. 
The heart of \GALA resides in Phase III, which coordinates a \textsf{Re‐ranking Agent} and a \textsf{Deep Dive Analysis Agent} in an iterative loop, updating the composite ranking \(R^{\mathrm{Curr}}\) until the \textsf{Re-ranking Agent} decides to conclude the analysis or max iteration is met. Finally, Phase IV’s \textsf{Remediation Agent} analyzes \(R^{\mathrm{Curr}}\)  and the LLM reasoning history into a structured incident summary accompanied by prioritized action recommendations. 

\subsection{Initial Hypothesis Generation}
\label{Initial Hypothesis Generation}
\subsubsection{Metrics-based Root Cause Ranking}

\GALA applies a metrics-based causal structure learning algorithm~\cite{pham2024baro,xin2023causalrca,lin2024root,shimizu2006lingam} to learn causal relationships represented as a DAG from the collected metrics data \(M\). This causal graph \(G_{M}\) serves as the foundation for identifying candidate root causes in complex microservice environments by modeling dependencies and potential failure propagation paths.
The framework then derives an initial metrics-based ranking by applying ranking algorithms (e.g., Random Walk~\cite{lawler2010random} or PageRank~\cite{lawler2010random}) to prioritize pods that exert the greatest causal influence on observed anomalies. These algorithms compute scores for each candidate service and generate a ranked list \(R^{\mathrm{Init}}_{M}\) that guides further investigation and focuses remediation efforts on the most probable sources of the incident.

However, purely metrics-based causal discovery exhibits two major shortcomings. First, causal structure learning often misidentifies edge directions and suffers from accuracy degradation without meticulous hyperparameter tuning or precisely known failure timestamps~\cite{pham2024root}. Second, these algorithms cannot inherently distinguish routine operational fluctuations from genuine anomalies, necessitating separate anomaly detection mechanisms to filter spurious signals~\cite{pham2024root}.
As demonstrated experimentally in Section~\ref{eval_rq1}, \GALA addresses these limitations by combining the initial metrics-based ranking with additional data modalities and LLM reasoning, enabling more granular and accurate analysis that leverages both causal discovery and semantic understanding from logs and traces.

\subsubsection{Trace-based Initial Root Cause Ranking}

As part of the \GALA framework, we introduce \TWIST, a method for computing an initial root cause ranking based on distributed trace analysis. Our novelty lies in integrating four established scores—\textsf{self-anomaly}, \textsf{trace impact}, \textsf{blast radius}, and \textsf{delay severity}—into a cohesive, quantitative root cause ranking framework. \TWIST applies statistical analysis to identify anomalous performance patterns and constructs a trace DAG to reveal dependencies, offering detailed insights into system behavior. Formally, the TWIST method computes:

\[
R^{\mathrm{Init}}_{C} \;=\; \TWIST(C),
\]

\noindent In conducting (1) span‐level anomaly detection via dynamic thresholding and DAG construction of each trace, and (2) service‐level scoring by integrating self‐anomaly, trace impact, blast radius, and delay severity into a composite score:
\[
\mathrm{score}(s) \;=\;\sum_{i=1}^4 w_i\,c_i(s),
\]
where each \(c_i(s)\in[0,1]\) quantifies one anomaly dimension and  \(w_i\) represents the corresponding weight.

\textbf{Trace Processing and Anomaly Detection.}
The initial phase of \TWIST comprises two principal operations. First, span-level anomaly detection is conducted via dynamic thresholding: each span's duration is compared against a threshold that adapts to historical performance patterns, isolating spans whose latencies significantly exceed expected norms. Second, we perform trace graph construction by modeling each distributed trace as a DAG, where nodes represent individual spans and edges encode parent-child call relationships. This DAG abstraction preserves the causal structure of request flows and enables analysis of how anomalies propagate through the service hierarchy.

\textbf{Service Ranking.}
To prioritize candidate services by their likelihood of fault origination, \TWIST computes the \(\mathrm{score}(s)\) for each service \(s\)  by integrating four normalized component scores. 
The {\textsf{Self-Anomaly Score ($c_1$)}} measures how frequently a service violates its own performance norms, computed as the proportion of \(s\)’s spans (or requests) that are flagged by span‐level anomaly detection. High values (close to 1) indicate that the service itself frequently exhibits latency or error deviations, consistent with SLO‐violation monitoring approaches~\cite{wu2020microrca}.
The \textsf{Trace Impact Score ($c_2$)} quantifies the involvement of the service \(s\) in all anomalous request traces, defined as the fraction of all anomalous traces in which \(s\) appears~\cite{chen2002pinpoint}. A larger value suggests that the service consistently participates in problematic executions.
The \textsf{Blast Radius Score ($c_3$)} captures a service's potential to propagate faults by measuring its average downstream fan-out in the trace DAG. Specifically, it calculates the mean number of immediate child spans invoked by each of \(s\)’s spans, aligning with observations that nodes with high degree contribute to widespread impact~\cite{wu2020microrca}. Higher values (closer to 1) of $c_3$ point to services that, when anomalous, could impact a larger portion of the system~\cite{wu2020microrca}.
The \textsf{Delay Severity Score ($c_4$)} emphasizes the magnitude of performance degradation experienced by \(s\), computed as the maximum latency excess of service \(s\)’s anomalous spans, normalized by the system‐wide maximum deviation. Values near unity highlight services that endure the most severe delays, as advocated in CIRCA~\cite{li2022causal}. This score near 1 indicates \(s\) experiences some of the most significant performance hit compared to other services, highlighting it as a strong underlying cause candidate.

These four component scores are then aggregated with a weighted average which yields a robust ordering of suspect services. Such weighted‐sum fusion has been shown to enhance diagnosis accuracy in prior frameworks, including MS‐Rank~\cite{ma2020self}, and mirrors other causality‐aware ranking methods that blend anomaly severity with graph‐topology insights~\cite{lin2024root,xing2025multi}. \TWIST distinguishes itself from other trace-based RCA methods by integrating multiple complementary scoring functions, each capturing a distinct dimension of service, and integrating them into an initial hypothesis that more comprehensively reflects the underlying cause.

\subsection{Pod-Centric Diagnostic Synthesis}
\label{Pod-Centric Diagnostic Synthesis}

Phase II of \GALA refines each candidate’s telemetry \((m_i, l_i, c_i)\) into three concise, pod-centric artifacts: (1) Temporal Performance Profiling, which visually expose temporal resource dynamics and anomaly patterns of \(m_i\); (2) Service Dependency Subgraph Extraction, which distill end-to-end call dependencies and predecessor/successor paths with \(c_i\); and (3) Error-Centric Log Abstraction, which filters and de-duplicates \(l_i\) to produce a crisp corpus of error events and exception signatures for each pod. By converting heterogeneous, high-volume data into these structured and visual formats, we present the LLM with a condensed diagnostic bundle \(D_i\) that enables deeper contextualization of performance degradations, causal bottlenecks, and failure signatures (Section~\ref{case_study}).

\subsubsection{Temporal Performance Profiling}

This module processes each pod’s raw metrics time‐series \(m_i\) to produce a concise, visual summary suitable for downstream LLM reasoning. Concretely, for each pod \(i\in\mathcal{P}\), the per‐second measurements in \(m_i\) are aggregated over non‐overlapping five‐minute intervals by computing summary statistics (e.g., mean, maximum, and percentile values) for each resource dimension (CPU, memory, I/O, etc.). These aggregated values are then rendered as a time‐series plot (PNG) and stored in the diagnostic bundle \(D_i\).
This five-minute aggregation strikes a balance between information preservation and efficiency: it reduces input volume to the LLM, prevents token/image-size constraints, and preserves temporal patterns necessary for effective RCA. The performance profiles complement execution trace graphs and error-centric logs by providing stable temporal context, enabling the \textsf{Deep Dive Agent} to conduct well-informed reasoning with multi-modal data.

\subsubsection{Service Dependency Subgraph Extraction}

All distributed traces \(C=\{c_i\}\) are first aggregated to construct a global service relationship graph, where nodes represent services and directed edges denote invocations. When the \textsf{Re-ranking Agent} requests context for pod \(i\), the \textsf{Deep Dive Agent} extracts the local subgraph \(g_{i}\) characterized by predecessor and successor sets. This Service Dependency Subgraph Extraction supports RCA by encoding causal pathways through which faults propagate, providing the LLM with structured graph context that improves hypothesis generation and accelerates convergence of the iterative reasoning loop.

\subsubsection{Error‐Centric Log Abstraction}

This module transforms each pod’s raw log stream \(l_i\) into a concise, error‐centric corpus, \(l_i^{\mathrm{ref}}\). Formally, we define  
\[
l_i^{\mathrm{ref}} \;=\;\mathrm{LogParser}(l_i),\quad l_i^{\mathrm{ref}}\subseteq l_i,
\]  
where \(\mathrm{LogParser}(\cdot)\) sorts entries by timestamp, filters out informational and debug lines by retaining only error-level severity or exception signatures, and eliminates semantic duplicates via message-level de-duplication. If error entries exceed a threshold, a representative random sample is selected; otherwise, the remainder is supplemented with non-error entries to help the LLM distinguish silent failures from log absence. The result, \(l_i^{\mathrm{ref}}\), is then incorporated into the diagnostic bundle \(D_i\).
This abstraction significantly increases signal-to-noise ratio, controls token budget consumption, and standardizes the log modality across pods. In the diagnostic bundle \(D_i\),  \(l_i^{\mathrm{ref}}\) complements temporal performance profiles and execution trace graphs by providing critical semantic evidence for richer, multi-modal causal inference.

\subsection{LLM Agentic Reasoning and Re‐ranking}
\label{LLM Agentic Reasoning and Re‐ranking}

In Phase III, \GALA orchestrates an iterative, multi‐agent ReAct~\cite{yao2023react} reasoning loop in which structured telemetry artifacts are analyzed, synthesized, and re‐evaluated by collaborating LLM agents. The goal is to progressively refine the root‐cause ranking until a the Re-ranking Agent concludes its analysis or max iterations are reached. This section details the agent architecture, prompt design, re‐ranking algorithm, and convergence criteria underpinning this process. The overall agentic workflow follows the process described in Figure~\ref{fig:GALA}; the LLM Agentic Reasoning and Re‐ranking frameworks consist of three distinct specialized agents: \textsf{Re-ranking Agent}, \textsf{Deep Dive Agent}, and \textsf{Remediation Agent}.

\subsubsection{Re-ranking Agent}

This agent acts as the main controller, directs the overall process, analyzes the deep dive diagnostic bundle \(D_i\), and determines next action iteratively. Given the initial rankings \(R^{Initial}_{M}\) and \(R^{Initial}_{C}\), this agent re-examines origin candidate rankings and decides on the next pod to analyze or finish the reasoning process. Each iteration begins with the Re‐ranking Agent ingesting the current ranking \(R^{\mathrm{Curr}}\), the predecessor/successor context from the trace subgraph, and the most recent summary, \(\sigma_i\), produced by the Deep Dive Agent for pod \(i\). Guided by a structured, chain‐of‐thought~\cite{wei2022chain} prompt, the Re‐ranking Agent re‐assesses all candidates, emits an updated ranking rendered in JSON, and decides whether to terminate ("\texttt{Finish}") or to request further analysis of a new candidate ("\texttt{Analyze Next}").

\begin{figure}[h]
    \centering
    \centerline{
        \includegraphics[width=0.42\textwidth]{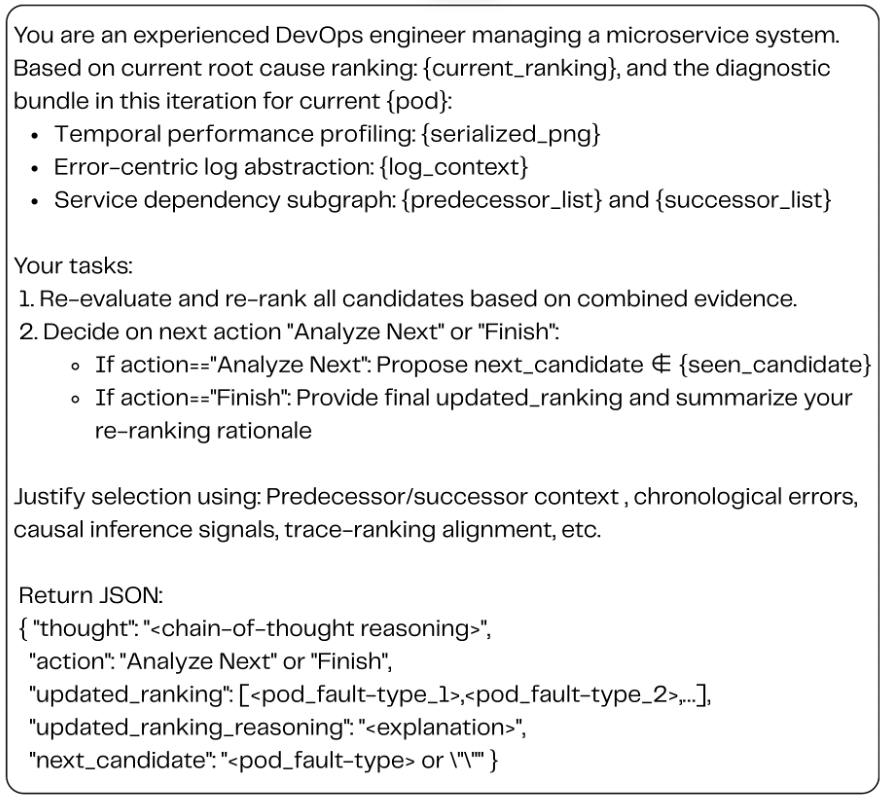}
    }
    \caption{The prompt template for \textsf{Re-ranking Agent}.} 
    \label{fig:prompt_template_reranking}
\end{figure}

The prompt template for Re-ranking Agent is shown in Figure ~\ref{fig:prompt_template_reranking}. This design leverages structured JSON output to ensure deterministic parsing and downstream automation. Empirical studies of chain‐of‐thought prompting demonstrate significant gains in multi‐step reasoning tasks~\cite{wei2022chain}, suggesting that explicit reasoning traces improve both ranking confidence and interpretability. As presented in Section~\ref{case_study}, this agent’s use of causal inference alignment conditions in the prompt further grounds its decisions in prior analysis, yielding more robust convergence than purely statistical ranking algorithms like BARO~\cite{pham2024baro}.

\subsubsection{Deep Dive Agent}

The Deep Dive Agent executes each “Analyze\_Next” request by retrieving the diagnostic bundle \(D_i\), which comprises the temporal performance profile, the extracted service dependency subgraph, and the error‐centric log abstraction for pod \(i\). Its prompt instructs the model to integrate these multi‐modal artifacts, Base64‐encoded metrics PNG plots, trace DAG context, and filtered logs, into a concise summary \(\sigma_i\) containing diagnostic observations, causal hypotheses, and reasoning pathway towards conclusion. Prior studies in multi‐modal LLM reasoning demonstrate that combining visual and textual evidence enhances interpretability and accuracy~\cite{alayrac2022flamingo}. The Deep Dive Agent’s output is appended to the interaction history and fed back to the Re‐ranking Agent.

\subsubsection{Remediation Agent} 

Once the Re‐ranking Agent issues “Finish,” the Remediation Agent aggregates the final ranking \(R^{\mathrm{Curr}}\) and the chronological chain of summaries \(\{\sigma_i\}\) to produce a narrative Incident Summary and a prioritized Action Recommendation checklist. This division of labor: Re‐ranking Agent for strategic origin candidate selection, Deep Dive Agent for focused evidence synthesis and Remediation Agent for textual output generation, ensures that \GALA converges on high‐confidence root‐cause hypotheses with minimal redundancy, while preserving a transparent reasoning trace for OCEs.

\medskip
\noindent\textbf{Prompt Design Considerations.}  
All three agents employ rigorously structured system-level instructions to elicit chain-of-thought reasoning~\cite{wei2022chain}, explicit JSON outputs, and consistent terminology. Key design elements include clear separation of duties, explicit enumeration of required fields, and fault-type definitions embedded in prompts to ensure reproducibility and minimize hallucination risk.

\subsection{Final Output Preparation}
\label{Final Output Preparation}

The Final Output Preparation of \GALA is tailored to alleviate information overload that OCEs face when confronted with large telemetry volumes, by coupling automated root cause identification with rich explanatory context and concrete remediation guidance. The output comprises:

\begin{enumerate}  [leftmargin=*]
    \item \textbf{A Final Root-Cause Ranking}, ordered by inferred fault likelihood and embedded with LLM-derived confidence annotations; 
    \item \textbf{An Incident Summary}, which reconstructs a narrative of the diagnostic reasoning—linking performance anomalies, causal-graph relationships, and error-log evidence to make the system’s reasoning process transparent; and
    \item  \textbf{A prioritized Action Recommendation list}, translating each identified cause into specific operational steps (e.g., configuration tweaks, resource scaling, dependency throttling) accompanied by estimated impact assessments. 
\end{enumerate}

By distilling excessive incident telemetries into a refined, structured report, \GALA reduces the troubleshooting workload of engineers, helps them rapidly pinpoint \textbf{what} went wrong, \textbf{why} it happened, and \textbf{how} to fix it. This integrated presentation ensures remediation decisions are both well-justified and immediately actionable, thereby shortening incident resolution time and enhancing overall system reliability.

\section{RCA Dataset Augmentation}

\subsection{Dataset Generation}
\label{RCA_dataset_generation}

Publicly available RCA datasets like RCAEval~\cite{pham2025benchmark} for microservice systems focus on ranking-based evaluation, providing incident scenarios annotated with correct root causes and requiring models to output ranked candidate lists. While this approach supports standard metrics such as  Top-k accuracy or mean reciprocal rank (MRR), recent studies~\cite{jiang2024xpert,kang2024quantitative,pei2025flow} show that practitioners prefer RCA systems that explain causal chains and provide actionable guidance, motivating the inclusion of high-quality explanations and recommendations in both dataset construction and evaluation. However, the data from these studies are often not open-sourced and cannot be used for future research.

To address these limitations, we introduce an LLM-based augmentation pipeline for open-source ranking-based RCA datasets like RCAEval~\cite{pham2025benchmark}, designed to produce high-quality reference summaries and prioritized, incident-specific recommendations. For each incident, we instantiate \GALA's \textsf{Remediation Agent} with the ground-truth root-cause label, alongside injected failure types, timestamp, and service name, alongside the microservice application information. The agent is guided by detailed annotation guidelines that emphasize technical precision, factual completeness, clarity, and actionability. Its outputs consist of a concise incident summary and three tailored remediation recommendations.
All generated artifacts undergo human expert review to ensure operational relevance and conformity with best practices. The evaluation panel comprised an AIOps Tech Lead with six years of combined research and industry experience, alongside two Senior Software Engineers, each with four years of on-call engineering experience. The resulting augmented dataset provides contextually grounded incident narratives and prescriptive benchmarks, supporting both LLM training and rigorous evaluation for practical RCA workflows.

\subsection{Human‐Guided LLM Evaluation}
\label{SURE_Score}

Standard automatic natural language generation evaluation metrics (e.g.,\ BLEU~\cite{papineni2002bleu}, ROUGE~\cite{lin2004rouge}, BERTScore~\cite{zhang2020bertscore}) quantify lexical or embedding overlap but fail to assess the logical coherence, actionability, and operational relevance that are essential in RCA. Human-guided LLM evaluation frameworks provide more reliable and contextually valid assessments than these automated metrics~\cite{chu2025thinkworkbettercombining}. Therefore, we introduce \surescore (\textbf{SU}mmarization \textbf{RE}commendation score), a human‐guided evaluation framework tailored to RCA tasks. SURE‐Score adapts the three‐stage structure of InteracEval~\cite{chu2025thinkworkbettercombining}: (1) Think‐Aloud elaboration of text attributes by both human experts and four different LLMs (\texttt{GPT-4.1}, \texttt{Gemini-2.5-Pro}, \texttt{Claude-Sonnet-4} and \texttt{Llama-4-Maverick}); (2) Semi‐automated clustering into domain‐specific evaluation components; and (3) Binary checklist‐based scoring. SURE‐Score differs fundamentally from InteracEval in its domain adaptation.  Whereas InteracEval’s checklist targets general NLG dimensions, SURE‐Score organizes its evaluation around four RCA‐critical dimensions (causal soundness, actionability, incident specificity and clarity; see Table~\ref{tab:SURE_score_dimensions}), ensuring that each checklist item probes evidence‐backed causal links, concrete remediation steps and incident‐focused detail, and incorporates expert‐validated prompts to enforce technical precision and operational feasibility.

\begin{table}[h]
\centering
\caption{SURE-Score Evaluation Dimensions.}
\label{tab:SURE_score_dimensions}
\small

\begin{tabularx}{\linewidth}{lX}
\hline
\textbf{Dimension} & \textbf{Description} \\
\hline
Causal Soundness & The logical coherence of the root cause narrative, requiring that each causal link is substantiated by evidence and that the explanation follows a justifiable progression from observed symptoms to root cause. \\
\hline
Actionability & The degree to which the recommendations are concrete, feasible, and directly address the identified root cause, penalizing suggestions that are vague or non-operational. \\
\hline
Incident Specificity & The extent to which both the summary and recommendations are concrete, detailed, and tightly focused on the specific incident, avoiding generic statements not pertinent to the case. \\
\hline
Clarity & The readability and unambiguity of the explanation and recommendations, with an emphasis on grammatical correctness and avoidance of excessive technical jargon. \\
\hline
\end{tabularx}
\end{table}

\surescore development proceeds via iterative human expert generation, LLM‐assisted clustering, and pilot annotation to guarantee coverage and eliminate ambiguity. During evaluation using \surescore, an LLM evaluator answers each Yes/No question in the human-LLM co-generated checklist, and dimension‐wise scores are computed as the proportion of positive responses (rescaled to a 1–5 scale). This protocol generates transparent, reproducible scores that closely align with operational practitioners’ judgments and highlights precise areas for model improvement.

\section{Evaluation}

In this section, we aim to evaluate \GALA by answering the following research questions (RQs):

\begin{itemize} [leftmargin=*]

\item \textbf{RQ1:} How effective is \GALA in RCA tasks?

\item \textbf{RQ2:} What is the contribution of each individual component of the \GALA framework to overall RCA performance?

\item \textbf{RQ3:} How do different LLMs affect the performance of the \GALA framework in RCA tasks?

\item \textbf{RQ4:} How effective is \GALA in incident summarization and action recommendation?

\end{itemize}

\subsection{Experiment Setup}

In this study, we describe the experimental setup used to evaluate \GALA, including dataset selection, LLM settings and VM specifications.

\textbf{Microservice Application.}
We choose the RE2-OnlineBoutique (OB) dataset from the RCAEval benchmark \cite{pham2025benchmark} to evaluate for RQ1-3, and augmented RE2-OB for RQ4 evaluation via the method proposed in Section~\ref{RCA_dataset_generation}. The OnlineBoutique application consists of 12 services that communicate using RPC. The RE2-OB dataset comprises 90 injected-fault scenarios spanning five services: \texttt{checkout service}, \texttt{currency service}, \texttt{email service}, \texttt{productcatalog service} and \texttt{recommendation service}. Six failure types (CPU hog, memory leak, disk I/O stress, network delay, socket errors and packet loss) provide a rigorous evaluation suite for RCA algorithms. Each service–fault combination was executed in three independent rounds, yielding 90 distinct incidents with detailed telemetry: per-pod metrics time-series, raw log streams, and distributed trace snapshots. 
For producing reference summaries and incident‐specific recommendations in dataset augmentation, we use \texttt{GPT-4.1-mini}, after human expert validation. The output is used as remediation ground truth for each incident.
For dataset augmentation, we use \texttt{GPT-4.1-mini} with human expert validation to produce reference summaries and incident-specific recommendations as remediation ground truth.

\textbf{LLM settings.}
To accommodate extensive telemetry and log context, we selected three GPT-family models: \texttt{GPT-4.1}, \texttt{GPT-4.1-mini}, and \texttt{GPT-4o-mini}, denoted as M1, M2, and M3, respectively. All models support context windows of at least 128K tokens and were invoked with a temperature of 1.0 to balance output diversity and coherence. The LLM reasoning and re-ranking agentic workflow reasoning was limited to six iterations per incident to control inference latency and computational cost.
For \surescore evaluation, we utilize \texttt{GPT-4.1-mini} as the primary model.

\textbf{Baseline Models.}
We compare GALA against three representative state-of-the-art RCA techniques: Granger~\cite{lin2024root}, CausalRCA~\cite{xin2023causalrca} and BARO~\cite{pham2024baro}, representing distinct strengths in temporal dependencies, structured causal models, and graph centrality analysis. We additionally report performance of PC~\cite{spirtes2001causation}, LiNGAM~\cite{shimizu2006lingam}, TraceRCA~\cite{li2021practical}, and Nezha~\cite{yu2023nezha} for comparative analysis. When root cause ranking requires causal graphs, we distinguish PageRank-based~\cite{bianchini2005inside} (PR) from random-walk-based~\cite{lawler2010random} (RW) variants.

\textbf{Implementation Detail.}
All experiments were executed on an Ubuntu 22.04 virtual machine with eight physical CPU cores, 16 GB RAM, and 500 GB SSD storage. The software stack was built on Python 3.10.12 without GPU acceleration, ensuring reproducibility under modest hardware constraints.

\subsection{Evaluation Metrics}

\textbf{Metrics for RQ1-3.} In order to capture both strict top‐1 precision and the broader ranking quality necessary for rapid incident triage, we measure these metrics following standard practice in the RCA literature~\cite{pham2025benchmark,wu2020microrca,lin2024root,pham2024baro,yu2023nezha,pham2024root}:
\begin{itemize} [leftmargin=*]
  \item \textbf{Accuracy@1} (Acc@1): Proportion of cases where the top‐1 ranked candidate is the true root cause.
  \item \textbf{Average Accuracy@k} (Avg@k): The mean fraction of true root causes appearing within the top-\(k\) positions, evaluated for \(k=3\) and \(k=5\).  
\end{itemize}

\noindent\textbf{Metrics for RQ4.}
We measure \GALA's incident summarization and action recommendation effectiveness with traditional NLG metrics (BLEU~\cite{papineni2002bleu},  ROUGE-L~\cite{lin2004rouge}, BERTScore~\cite{zhang2020bertscore},  Jaccard Similarity~\cite{jaccard1901distribution}, Cosine Similarity~\cite{salton1986introduction}) and our \surescore (Section~\ref{SURE_Score}).

\subsection{Evaluation Results}

\subsubsection{RQ1: Effectiveness of \GALA}

\label{eval_rq1}

Table~\ref{tab:eval-summary} presents AC@1, Avg@3 and Avg@5 for nine baseline models, and compares with \GALA's performance when using Granger‐PR, CausalRCA and BARO as the initial metrics-based root cause ranking methods. We compare two \GALA variants with different LLM implementations (M1 and M2) against each baseline method.

\begin{table}[h]
\centering
\small
\renewcommand{\arraystretch}{0.73}
\caption{Evaluation results (AC@1, Avg@3, Avg@5) in percentage for all methods and variants on RE2‐OB.}
\label{tab:eval-summary}
\begin{tabular}{lrrr}
\toprule
\textbf{Method}             & \textbf{AC@1} & \textbf{Avg@3} & \textbf{Avg@5} \\
\midrule
Granger-RW                     &  3.33         & 25.19          & 39.33          \\
LiNGAM-RW                      &  3.33         & 24.81          & 39.33          \\
PC-PR                          & 15.56         & 25.56          & 38.44          \\
PC-RW                          &  3.33         & 24.81          & 38.67          \\
TraceRCA                       & 13.19         & 41.39          & 61.32          \\
Nezha                          &  8.89         & 12.59          & 15.56          \\

\midrule

Granger-PR                     & 12.22         & 24.07          & 36.67          \\
GALA (Granger-PR, M1)           & 12.22         & 24.81 $\uparrow$         & 39.11 $\uparrow$         \\
GALA (Granger-PR, M2)           & 15.56 $\uparrow$        & 34.07 $\uparrow$         & 44.89 $\uparrow$         \\

\midrule
CausalRCA                      & 26.67         & 47.78          & 60.44          \\
GALA (CausalRCA, M1)            & 41.11 $\uparrow$        & 59.63 $\uparrow$         & 69.11 $\uparrow$         \\
GALA (CausalRCA, M2)            & 31.11 $\uparrow$       & 52.22 $\uparrow$       & 64.22 $\uparrow$        \\

\midrule
BARO                           & 14.44         & 61.48          & 74.22          \\
GALA (BARO, M1)                 & 38.89 $\uparrow$        & \textbf{68.89 $\uparrow$}        & \textbf{79.56 $\uparrow$}       \\
GALA (BARO, M2)                 & \textbf{42.22 $\uparrow$}      & 67.78 $\uparrow$         & 77.33 $\uparrow$         \\

\bottomrule
\end{tabular}
\end{table}

\begin{table*}[ht!]
\centering
\small
\setlength{\tabcolsep}{1pt}
\renewcommand{\arraystretch}{0.82}
\caption{Per‐fault evaluation result (AC@1, Avg@3, Avg@5) on the RE2‐OB benchmark. \textbf{Bold} indicates the best performance of that measurement; \underline{underlined} entries denote \GALA ties with the baseline method.}
\label{tab:per_fault_results}
\resizebox{\textwidth}{!}{%
\begin{tabular}{l*{18}{r}}
\toprule
\textbf{Method}
  & \multicolumn{3}{c}{\textbf{CPU}}
  & \multicolumn{3}{c}{\textbf{Delay}}
  & \multicolumn{3}{c}{\textbf{Disk}}
  & \multicolumn{3}{c}{\textbf{Loss}}
  & \multicolumn{3}{c}{\textbf{Mem}}
  & \multicolumn{3}{c}{\textbf{Socket}} \\
\cmidrule(lr){2-4}
\cmidrule(lr){5-7}
\cmidrule(lr){8-10}
\cmidrule(lr){11-13}
\cmidrule(lr){14-16}
\cmidrule(lr){17-19}
  & AC@1 & Avg@3 & Avg@5 
  & AC@1 & Avg@3 & Avg@5 
  & AC@1 & Avg@3 & Avg@5 
  & AC@1 & Avg@3 & Avg@5 
  & AC@1 & Avg@3 & Avg@5 
  & AC@1 & Avg@3 & Avg@5 \\
\midrule
Granger-RW &  6.67 & 35.56 & 48.00   &  0.00 & 24.44 & 37.33   &  6.67 & 33.33 & 42.67   &  0.00 &  8.89 & 28.00   &  6.67 & 24.44 & 37.33   &  0.00 & 24.44 & 42.67 \\
LiNGAM-RW  &  6.67 & 33.33 & 45.33   &  0.00 & 24.44 & 38.67   &  6.67 & 33.33 & 44.00   &  0.00 &  8.89 & 28.00   &  6.67 & 24.44 & 37.33   &  0.00 & 24.44 & 42.67 \\
PC-PR        & 13.33 & 24.44 & 37.33   & \underline{26.67} & 31.11 & 44.00   &  6.67 & 15.56 & 32.00  & 0.00 & 15.56 & 25.33 & 20.00   & 31.11 & 41.33 & \underline{26.67}   & 35.56 & 50.67  \\
PC-RW      &  6.67 & 35.56 & 48.00   &  0.00 & 24.44 & 37.33   &  6.67 & 31.11 & 38.67   &  0.00 &  8.89 & 28.00   &  6.67 & 24.44 & 37.33   &  0.00 & 24.44 & 42.67 \\
TraceRCA            & 25.00 & 62.50 & 76.25   & 20.00 & 33.33 & 52.00   & 26.67 & 51.11 & 70.67   &  6.67 & 31.11 & 48.00   & 20.00 & 31.11 & 58.67   &  0.00 & 37.78 & 61.33 \\
Granger-PR   & 20.00 & 31.11 & 44.00   &  4.44 & 17.33 & 20.00   & 26.67 & 55.56 & 65.33   &  6.67 & 22.22 & 36.00   & 20.00 & 37.78 & 52.00   &  6.67 & 22.22 & 34.67 \\
CausalRCA           & 20.00 & 33.33 & 44.00   & 20.00 & 53.33 & 65.33   & 26.67 & 55.56 & 65.33   & 26.67 & 42.22 & 56.00   & 53.33 & 64.44 & 74.67   & 13.33 & 37.78 & 57.33 \\
BARO                &  0.00 & 51.11 & 64.00   &  0.00 & 48.89 & 66.67   & 13.33 & 68.89 & 81.33   & \underline{40.00} & 68.89 & 80.00   & 40.00 & 68.89 & 86.67   &  0.00 & 53.33 & 66.67 \\
\midrule
GALA (BARO, M1)     &\textbf{46.67}	&\textbf{68.89}	&\textbf{77.33}	&\underline{26.67}	&\textbf{57.78}	&\textbf{70.67}	&\textbf{53.33}	&\textbf{77.78}	&\textbf{86.67}	&\underline{40.00}	&\textbf{71.11}	&\textbf{82.67}	&46.67	&82.22	&89.33	&20.00	&55.56	&70.67 \\
GALA (BARO, M2)      & 40.00 & \textbf{68.89} & \textbf{77.33}   & 13.33 & 55.56 & 68.00   & \textbf{53.33} & 73.33 & 84.00   & \underline{40.00} & 60.00 & 72.00   & \textbf{80.00} & \textbf{91.11} & \textbf{94.67}   & \underline{26.67} & \textbf{57.78} & \textbf{68.00} \\

\bottomrule
\end{tabular}%
}
\end{table*}

The most significant performance boost occurs for \GALA (BARO, M2), where Top‐1 accuracy reaches 42.22\%, Avg@3 67.78\% and Avg@5 77.33\%, respectively, and for GALA (CausalRCA, M1), which attains the highest Avg@3 (59.63\%) and Avg@5 (69.11\%) among all variants. With using Granger‐PR as initial metrics-based ranking, \GALA (M1) preserves top‐1 accuracy while boosting Avg@3 by +0.74\% and Avg@5 by +2.44\%, and \GALA(M2) further increases AC@1 by +3.34\%, Avg@3 by +10.00\% and Avg@5 by +8.22\%. On CausalRCA, \GALA (M1) yields gains of +14.44\% in AC@1, +11.85\% in Avg@3 and +8.67\% in Avg@5, with \GALA (M2) still delivering improvements of +4.44\% across all three metrics.

These results confirm that \GALA's initial hypothesis generation efficiently narrows down candidates, and the subsequent LLM Agentic Reasoning modules effectively improve original rankings. \GALA's agentic, multi-modal reasoning architecture iteratively combines subtle fault signatures (e.g., memory-leak spikes aligned with specific log entries) and leverages trace subgraphs for attributing downstream performance impacts.

Table 4 breaks down performance across six fault categories. For all measurements and fault categories, GALA achieves best or tied performance compared to baselines. Memory faults exhibit the most dramatic improvements: GALA (BARO,M2) attains 80.00\% AC@1 and 91.11\% Avg@3, more than doubling BARO's 40.00\%. Socket faults increase from 0\% to 20.00\% (M1) and 26.67\% (M2). Average-rank metrics follow similar trends, with GALA (BARO,M2) reaching 94.67\% Avg@5.

We logged LLM-agent interaction rounds per experiment (capped at 6) for each fault category, resulting in an overall mean of 5.54 rounds. Memory faults required the deepest reasoning (5.93 rounds), followed by CPU (5.80) and packet-loss (5.60), whereas delay anomalies converged most rapidly (5.00 rounds). This per-fault iteration depth mirrors our AC@1 improvements: categories with largest accuracy gains (\texttt{Mem} and \texttt{CPU}) correspond to longer reasoning loops, suggesting \GALA dynamically allocates more agentic deliberation when diagnosing complex, resource-intensive failures.

In contrast, traditional methods operate on single modalities or static heuristics: Granger-PR's pairwise correlations risk misattribution; CausalRCA's global causal inference omits log semantics; BARO's pagerank centrality ignores fine-grained semantics. None perform iterative re-ranking, leading them to overlook root causes with weak metric signals but strong semantic cues.

\noindent\textbf{RQ1 Key Takeaway.}  
\GALA's agentic, multi‐modal reasoning significantly enhances RCA performance by seeding candidate hypotheses and iteratively refining them via LLM‐driven re‐ranking over structured diagnostic bundles.

\subsubsection{RQ2: Ablation Study}

\begin{figure}[h]
  \centering
  \begin{subfigure}[b]{0.515\linewidth}
    \centering
    \includegraphics[width=\linewidth]{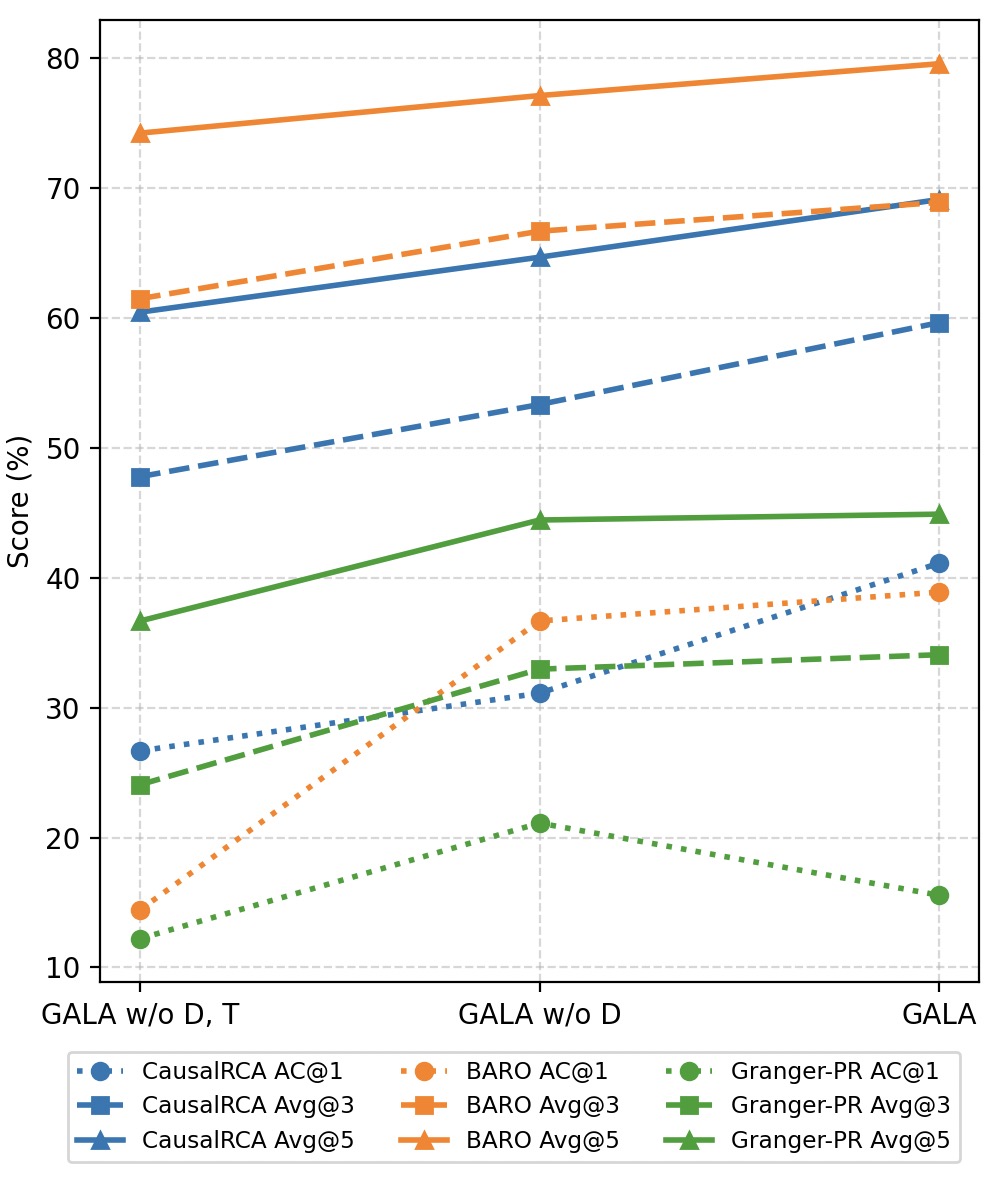}
    \caption{Ablation study.}
    \label{fig:rq2-A}
  \end{subfigure}\hfill
  \begin{subfigure}[b]{0.48\linewidth}
    \centering
    \includegraphics[width=\linewidth]{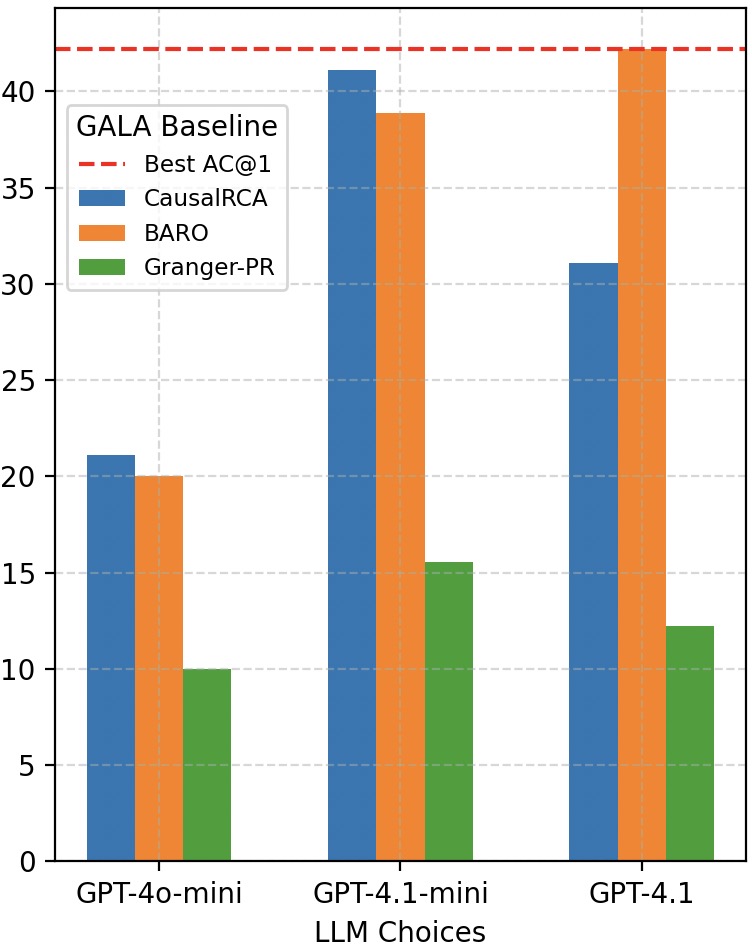}
    \caption{Impact of LLM.}
    \label{fig:rq3-B}
  \end{subfigure}
  \caption{Left figure shows the ablation study result of GALA w/o components using CausalRCA, BARO and Granger-PR as initial ranking methods. Right figure presents the impact of LLM Model Choice on AC@1 for GALA. }
  \label{fig:rq-2-3-plots}
\end{figure}

Figure~\ref{fig:rq2-A} presents an ablation study of the same underlying LLM model configured with: (i) \textbf{\GALA w/o Deep Dive (D), \TWIST (T):} Only metrics based ranking without initial hypothesis and the deep dive analysis, (ii) \textbf{\GALA w/o D:} \GALA adding \TWIST and not involving iterative deep dive, and (iii) \textbf{\GALA:} The full \GALA framework.

The “\GALA w/o D, T” variant's low AC@1 and average‐rank scores confirm the limitations of baseline metrics-based methods. Introducing the \TWIST module alone (\emph{w/o D}) outputs a consistent, mid‐level uplift across all baselines; for example AC@1 rises from 26.7\% to 31.1\% on CausalRCA and from 14.4\% to 36.7\% on BARO, demonstrating that global telemetry‐driven re‐scoring significantly improves candidate ordering in the initial hypothesis generation.

Adding the Deep Dive Agent on top of TWIST delivers further gains, as evidenced by the full \GALA curves. The incremental margin between "\GALA w/o D" and full \GALA confirms that the Deep Dive component's detailed multi-modal reasoning—combining time-series anomaly inspection, semantic log summarization, and trace-subgraph analysis—provides fine-grained re-ranking that maximizes AC@1, Avg@3 and Avg@5. These results substantiate that both modules contribute distinct and complementary benefits: \TWIST achieves broad performance lifts via initial hypothesis generation, while Deep Dive refines top-ranked candidates through contextual, evidence-based analysis.

\noindent\textbf{RQ2 Key Takeaway.}  
Integrating the \TWIST module in \GALA yields substantial ranking performance gains, and the subsequent addition of the agentic workflow with Deep Dive Agent further refines candidate prioritization.

\subsubsection{RQ3: Impact of LLM Selection on GALA Performance}

Figure~\ref{fig:rq3-B} isolates AC@1 to evaluate the effect of LLM "intelligence" within the \GALA framework. By holding the underlying re-ranking logic constant and varying only the LLM choice (GPT-4o-mini, GPT-4.1-mini, GPT-4.1), we observe a clear monotonic trend: GPT-4.1 achieves the highest AC@1 across all three baselines, confirming that more capable LLMs yield superior fault-localization accuracy. Notably, GPT-4.1-mini attains near-peak performance, achieving only marginally below full GPT-4.1, suggesting it as a cost-effective alternative when balancing inference expense against diagnostic quality. In contrast, GPT-4o-mini consistently under-performs, underscoring the sensitivity of GALA's ranking efficacy to the underlying LLM's reasoning capacity.

The magnitude of AC@1 improvement varies by baseline: the BARO foundation exhibits the largest gains (up to 42.22\%), while enhancements on Granger-PR remain more modest, indicating that the choice of causal inference or graph-based precursor interacts with LLM prowess to determine final accuracy. These results demonstrate that upgrading to a stronger LLM systematically elevates GALA's root-cause localization, with GPT-4.1-mini offering a compelling trade-off for real-world deployment.

\noindent\textbf{RQ3 Key Takeaway.}
 The AC@1 performance of \GALA increases monotonically with LLM capability. GPT-4.1 achieves the highest gains while GPT-4.1-mini delivers near-peak, cost-effective performance.

\begin{figure*}[h!]
    \centering
    \centerline{
        \includegraphics[width=\textwidth]{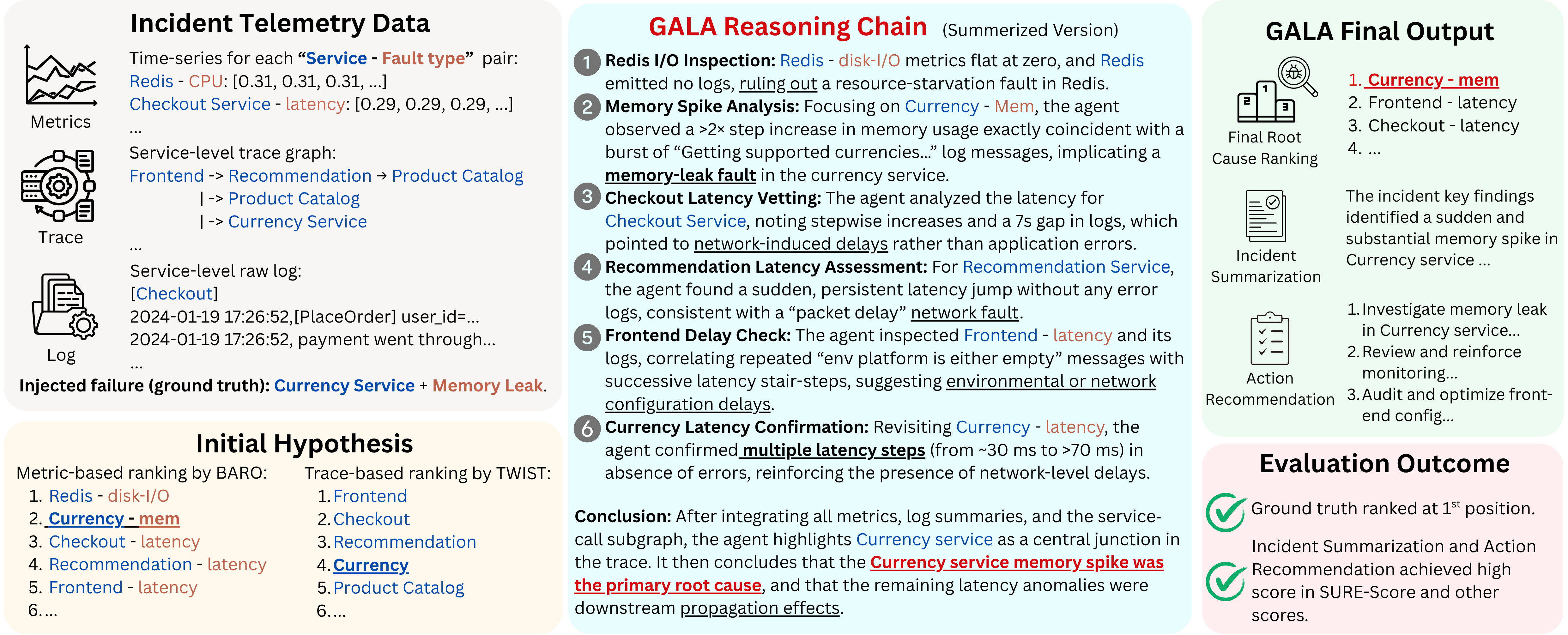}
    }
    \caption{A Case Study presenting \GALA's reasoning chain. In this case \GALA successfully improved the ranking to the top-1 position. The incident summarization and recommendation also achieved optimal score.}

    \label{fig:case_study}
\end{figure*}

\subsubsection{RQ4: How effective is \GALA in incident summarization and action recommendation?}

\begin{table}[ht]
\centering
\caption{LLM Summarization and Recommendation Evaluation with best results highlighted in bold.}
\label{tab:rq4_evaluation}
\small
\renewcommand{\arraystretch}{0.75}

\begin{tabular}{@{}l l r r@{}}
\toprule
\textbf{Category}         & \textbf{Metric}              & \textbf{Baseline} & \textbf{GALA} \\
\midrule
\multirow{5}{*}{Traditional}
  & BLEU                         & 0.0090            & 0.0068          \\
  & ROUGE-L                      & 0.1128            & 0.1023          \\
  & BERTScore                    & –0.0224           & –0.0327         \\
  & Cosine Similarity            & 0.2680            & 0.2775          \\
  & Jaccard                      & 0.0789            & 0.0727          \\
\midrule
\multirow{5}{*}{SURE‐Score}
  & Causal Soundness             & 3.41              & \textbf{4.42}            \\
  & Actionability                & 4.96              & 4.98            \\
  & Clarity                      & 3.65              & 3.55            \\
  & Incident Specificity         & 3.10              & \textbf{4.30}            \\
  & \textbf{Average}             & 3.78              & \textbf{4.31}            \\
\bottomrule
\end{tabular}
\end{table}

The results in Table~\ref{tab:rq4_evaluation} indicate that \GALA's LLM-driven summarization and recommendation produces comparable scores to the baseline on traditional n-gram overlap metrics (BLEU, ROUGE-L, Jaccard) and even outperforms it slightly in semantic similarity (CosineSimilarity), though it underperforms on BERTScore. The comparable performance can be partly attributed to their use of identical prompt templates (Section~\ref{RCA_dataset_generation}), which yields similar lexical choices and n-gram distributions. Moreover, these traditional overlap and embedding-based metrics are largely insensitive to semantic validity, focusing on surface form rather than factual accuracy or actionable quality.

Recognizing that traditional scores inadequately reflect the true operational effectiveness of incident summaries and remediation suggestions, we evaluate \GALA using \surescore. The second half of Table~\ref{tab:rq4_evaluation} presents mean ratings across four human-centric dimensions. \GALA achieves substantial gains in \textsf{Causal Soundness} (4.42 vs. 3.41) and \textsf{Incident Specificity} (4.30 vs. 3.10), demonstrating its unique ability to fuse causal inference outputs with LLM-based reasoning to produce logically coherent and highly targeted diagnostics. The overall \surescore average rises to 4.31, underscoring the system's strength in delivering precise, actionable root-cause explanations.
In contrast, differences in \textsf{Actionability} (4.98 vs. 4.96) and \textsf{Clarity} (3.55 vs. 3.65) are marginal, indicating that while \GALA's advanced reasoning markedly improves causal rigor and specificity, it retains comparable usability and readability relative to the baseline approach.

\noindent \textbf{RQ4 Key Takeaway:} Evaluating RCA-specific NLG tasks with traditional NLG metrics is not effective. \GALA's integration of causal inference outputs and LLM-based reasoning yields markedly higher generation performance when evaluating with \textit{SURE‐Score}.

\section{Case Study}

\label{case_study}

To illustrate the practical benefits of \GALA's multi‐modal, agentic reasoning, we present in Figure~\ref{fig:case_study} a detailed case study involving a synthetic memory‐leak fault injected into the \texttt{Currency} Service. Initially, the metrics‐based BARO approach incorrectly identified \texttt{Redis} disk-I/O as the primary root cause, while the trace‐based \TWIST method prioritized front‐end components. In contrast, \GALA’s LLM‐driven reasoning loops systematically excluded \texttt{Redis} through analysis of stable disk‐I/O metrics and the absence of relevant log events. \GALA then integrated multimodal evidence, including memory usage anomalies and latency issues in the \texttt{Checkout}, \texttt{Recommendation}, and \texttt{Front‐end} services, together with structural insights from the service‐call graph, pinpointing the \texttt{Currency} service as a critical junction. Through this iterative synthesis, \GALA accurately identified the \texttt{Currency} service memory leak as the primary root cause, enabling precise recommendations for enhanced memory monitoring and configuration auditing.  

\section{Related Work}

\noindent\textbf{LLM for RCA.}
The application of LLM to automate RCA tasks in microservice systems has emerged as an active research area, with approaches broadly categorized into three paradigms~\cite{pang2024large,pang2024hybrid,xing2023fusion,zhuang2022gan,saha2022mining,chakraborty2023esro,he2023construction,kuang2024knowledge,zhang2024automated,roy2024exploring,ahmed2023recommending,wang2024rcagent,xie2024cloud,chen2024automatic,zhang2024lm,gendron2024counterfactual,pei2025flow,xuopenrca,lewis2020retrieval,wang2024large,jiang2024xpert,li2024realtcd}. 
Fine-tuning approaches~\cite{saha2022mining,ahmed2023recommending,chen2024automatic,pang2024large,pang2024hybrid,xing2023fusion} adapt pre-trained models to domain-specific incident characteristics, enabling specialized classification and summarization of complex failure events.
Embedding-based techniques~\cite{zhuang2022gan,saha2022mining,chakraborty2023esro,he2023construction,kuang2024knowledge} leverage rich vector representations from models like BERT or specialized encoders to capture semantic and contextual nuances in telemetry data. 
Third, prompt-based strategies utilize in-context learning~\cite{zhang2024automated,roy2024exploring}, chain-of-thought reasoning~\cite{wei2022chain,ahmed2023recommending,wang2024rcagent,xie2024cloud,chen2024automatic,zhang2024automated,roy2024exploring,zhang2024lm,gendron2024counterfactual,pei2025flow,xuopenrca}, and retrieval-augmented generation~\cite{lewis2020retrieval,wang2024large,jiang2024xpert,li2024realtcd} to dynamically guide LLM outputs and synthesize diverse information sources for enhanced causal inference. 
Among existing work, OpenRCA~\cite{xuopenrca} pioneered an agentic workflow using cooperative LLM agents but is limited to code-generation tasks with uniform telemetry sampling, achieving only 11\% case-resolution rates. \GALA differs by employing specialized preprocessing pipelines for each telemetry modality and orchestrating iterative multi-modal reasoning that enhances fault localization while generating comprehensive incident summaries and remediation recommendations.

\noindent\textbf{Traditional and Multi-modal RCA.} 
Classical RCA approaches rely on statistical techniques and graph-based analysis~\cite{lin2024root,xin2023causalrca,shimizu2006lingam,spirtes2001causation,pham2024baro,wu2020microrca,chen2022deep,xie2023point,li2021practical}. Causal inference methods use Granger causality~\cite{lin2024root} and structural equation modeling~\cite{xin2023causalrca,shimizu2006lingam,spirtes2001causation} to identify relationships in metrics time series. Graph-based approaches~\cite{pham2024baro,wu2020microrca} leverage service topology with centrality measures and ranking algorithms~\cite{lawler2010random,bianchini2005inside}. However, these single-modality methods suffer from key limitations: inability to incorporate semantic information from logs, sensitivity to hyperparameter tuning, and lack of explanatory context.
Recent multi-modal approaches~\cite{pham2024baro,tao2024giving,sun2025interpretable,liu2022microcbr,zheng2024multi,yu2023nezha,fu2025msfsanomaly} address these limitations by integrating heterogeneous observability data. Unified representation methods like Nezha~\cite{yu2023nezha} consolidate metrics, logs, and traces as event graphs, while MSoFSAnomaly~\cite{fu2025msfsanomaly} selects fault-sensitive features to reduce redundancy. Structural modeling approaches such as MULAN~\cite{zheng2024multi} use contrastive learning and attention mechanisms for multi-modal fusion.
\GALA fundamentally differs by preserving native characteristics of each telemetry modality within structured diagnostic bundles rather than collapsing them into unified embedding spaces. This design combines statistical rigor with LLM-driven semantic reasoning, enabling richer multi-modal reasoning while maintaining interpretability and specificity of individual data sources for more robust and actionable root cause analysis.

\section{Threats to Validity.}

\textbf{Internal Threats.} 
Several factors may influence our findings' reliability. The choice of specific LLM models and configurations affects reasoning capability, context window size, and prompt sensitivity. Non-deterministic parameters like temperature introduce variability, particularly in multi-agent workflows relying on iterative prompts. To mitigate these threats, we selected a limited set of LLM models with standardized, carefully validated prompts. API-level changes in LLM services could introduce unexpected behavior changes. Human-guided evaluations by domain expert co-authors might introduce evaluator bias; we mitigated this through standardized rubrics, multiple evaluators, and strict consistency protocols.

\noindent\textbf{External Threats.} 
While our evaluation used publicly available RCA datasets such as RCAEval~\cite{pham2025benchmark}, the generalizability of \GALA might be limited by the specific nature of these datasets. Our evaluation scope focuses on particular telemetry types (e.g., CPU usage, latency), incident patterns, and service architectures, potentially differing from real-world production environments. Moreover, LLM agents are sensitive to input data formats, and their performance could degrade with unfamiliar or complex telemetry structures. Although our framework is modular, adapting \GALA effectively across diverse telemetry types and novel incident scenarios requires additional prompt refinement, and validation, which presents opportunities for future research.

\section{Conclusions}

This paper introduces \GALA, a novel framework that addresses the critical challenges of RCA in microservice systems by integrating multi-modal telemetry data through an LLM agentic workflow. \GALA's four-phase architecture—combining dual-modality hypothesis generation via metrics-based causal inference and the novel TWIST trace-based scoring module, structured diagnostic bundle synthesis, iterative LLM-driven reasoning and re-ranking, and comprehensive output generation—significantly outperforms existing approaches across multiple evaluation metrics. Our experimental results on the RCAEval benchmark demonstrate substantial improvements, with accuracy@1 reaching 42.22\% compared to 14.44\% for the baseline, while our \surescore evaluation framework reveals superior performance in causal soundness and incident specificity. By bridging the gap between fault identification and practical remediation through automated generation of incident summaries and actionable recommendations, \GALA represents a significant advancement toward intelligent, interpretable RCA that can effectively support on-call engineers. The future work includes exploring integration with even more data modalities, knowledge graphs, and human-in-the-loop refinement for enhanced deployment scalability.

\bibliographystyle{ACM-Reference-Format}
\bibliography{gl-rca}

\appendix

\end{document}